\documentclass[10pt,twocolumn,letterpaper]{article}

\usepackage{cvpr}
\usepackage{times}
\usepackage{epsfig}
\usepackage{graphicx}
\usepackage{amsmath}
\usepackage{amssymb}
\usepackage{comment}
\usepackage{algorithm}
\usepackage{algorithmic}
\usepackage{multirow}
\usepackage{subfig}
\usepackage{float}

\usepackage[breaklinks=true,bookmarks=false]{hyperref}
\usepackage{color}
\usepackage{xcolor}
\usepackage[normalem]{ulem}

\usepackage[breaklinks=true,bookmarks=false]{hyperref}

\cvprfinalcopy 

\def\cvprPaperID{****} 

\begin{document}

\title{Mixture of Bilateral-Projection Two-dimensional Probabilistic Principal Component Analysis}

\author{Fujiao Ju$^1$, Yanfeng Sun$^1$, Junbin Gao$^2$, Simeng Liu$^1$ and Yongli Hu$^1$\\
{\small $^1$College of Metropolitan Transportation,
Beijing University of Technology, Beijing 100124, China}\\
{\small School of Computing and Mathematics, Charles Sturt University, Bathurst, NSW 2795, Australia}\\
{\tt\small \{jufujiao2013,smliu\}@emails.bjut.edu.cn; \{yfsun,huyongli\}@bjut.edu.cn; jbgao@csu.edu.au}
}

\maketitle

\begin{abstract}
The probabilistic principal component analysis (PPCA) is built upon a global linear mapping, with which it is insufficient to model complex data variation. This paper proposes a mixture of bilateral-projection probabilistic principal component analysis model (mixB2DPPCA) on 2D data. With multi-components in the mixture, this model can be seen as a `soft' cluster algorithm and has capability of modeling data with complex structures. A Bayesian inference scheme has been proposed based on the variational EM (Expectation-Maximization) approach for learning model parameters.   Experiments on some publicly available databases show that the performance of mixB2DPPCA has been largely improved, resulting in more accurate reconstruction errors and recognition rates than the existing PCA-based algorithms. 
\end{abstract}

\section{Introduction}

Principle Component Analysis (PCA) \cite{Bishop2006} is one of  popular dimensionality reduction methods widely used in image analysis \cite{HoyerHyvarinen2000, KeSukthankar2004}, pattern recognition \cite{HeYanHuNiyogiZhang2005, LuPlataniotisVenetsanopoulosLi2006} and machine learning \cite{KriegelKrogerSchubert2008} for data analysis. It can be  derived under algebraic framework. However, algebraic models don't have flexibility of providing confidence information of the model when dealing with noisy data. This is due to the absence of an associated probability density or generative model in algebraic framework. 

To compensate the algebraic PCA drawbacks, Tipping and Bishop \cite{TippingBishop1999} firstly proposed a probabilistic PCA model, called PPCA. Under the probabilistic framework, PPCA takes advantage of Bayesian learning and inference by combining the likelihood with appropriate priors. As a result, the observed data are regarded as random variables, generated from a set of latent random variables which follow the Gaussian distribution of zero mean and identity covariance, with additive noises following a Gaussian distribution with zero mean and an isotropic covariance.
Under such a probabilistic learning framework, the model parameters in PPCA can be easily solved by the maximum likelihood estimation (MLE). Much progress has been made based on PCA and PPCA in the last couple of decades \cite{ArchambeauDelannayVerleysen2006,Gao2007}. 

PPCA and standard PCA methods can be interpreted in many ways, one of which assumes that the observed high-dimensional data are generated from their low-dimensional factors through a linear model with the corruption of Gaussian noise. So those algorithms essentially use a linear model for representing the entire data in a low dimensional subspace. It may be insufficient to model data with large variation caused by, for example, pose, expression and lighting in face recognition. Thus the application scope of PPCA and PCA-based methods is necessarily somewhat limited by its global linearity assumption. An alternative improving paradigm is to model the complex manifold with a mixture of local linear PPCA sub-models. Thus the single PCA model could be extended to a mixture of such sub-models.

A number of `mixture of PPCA' have been proposed in literature. The first work was done by Ghahramani and Hinton \cite{GhahramaniHinton1996}. They presented an exact Expectation-Maximization (EM) algorithm for fitting the parameters of the mixture of factor analyzers. By constraining the error covariance to be a diagonal matrix whose elements are usually equal, the mixture of factor analyzers became the mixture of PPCA \cite{TippingBishop1999a}. Bishop and Tipping \cite{BishopTipping1998} extended the mixture of PPCA model to achieve a hierarchical mixture model. Su and Dy \cite{SuDy2004} introduced an automated hierarchical mixture of PPCA algorithm, which utilizes the integrated classification likelihood as a criterion for splitting and stopping the addition of hierarchical levels. Kim \emph{et al}. \cite{KimKimBang2003} proposed a fast and sub-optimal selection method of model order such as the number of mixture components and the number of PCA bases for the PCA mixture model, consisting of a combination of many PCAs. In addition, under the assumption of the Student-$t$ distribution, the related research includes the mixture model of Student-$t$ components \cite{PeelMcLachlan2000}, which actually is a generalized mixture of Gaussian model without considering subspace structures, and more recent work such as the robust subspace mixture model \cite{RidderFranc2003}, in which both the likelihood and the latent variables were supposed to follow the Student-$t$ distribution and the EM algorithm was applied to the model. In 2005, Archambeau \cite{Archambeau2005} discussed the robust models in the context of finite mixture models, and a similar work for the mixture of the robust Laplacians was presented in \cite{GaoXu2007}. These mixture models are important as it enables one to model nonlinear relationships by aligning a collection of such local models.

The aforementioned models are concerned with vectorial data. 
In order to apply these methods to 2D data, a typical workaround way is to vectorize 2D data. Vectorizing 2D data not only results in very high-dimensional data, causing the problem of the curse of dimensionality \cite{XieYanKwokHuang2008}, but also ignores valuable information on the spatial relationship among 2D data. Instead of using vectorization, PCA approaches for two-dimensional data (2DPCA) have been proposed \cite{WangWang2013,YangZhangFrangiYang2004,YuBiYe2008}, 
to generally extract features of 2D data under the assumption of Gaussian noises. Ju \emph{et al}. \cite{JuSunGaoHuYin2015} proposed a probabilistic 2DPCA model to deal with outlier noises by using Laplacian distribution. This model benefits outlier detection. Wang \emph{et al}. \cite{WangChenHuLuo2008} extended the probabilistic 2DPCA to a mixture of local probabilistic 2DPCA models (MP2DPCA). MP2DPCA offers a tempting prospect of being able to model data with complex variation.

MP2DPCA model regards each row vector of the 2D data as a observed sample and used all rows to train the mixture model, resulting in mean vectors from the mixture model. This is essentially a unilateral projection based scheme, where only one side multiplication is taken into account. The unilateral scheme usually preserves the correlation information among the row/column vectors of the images and more parameters are needed to well represent an image. To tackle these problems, a bilateral-projection scheme is favored. In this study, our intention is propose a mixture of bilateral-projection-based probabilistic 2DPCA (mixB2DPPCA) model. Different from MP2DPCA, we regard each 2D images as observed samples in their natural shape and reduce 2D dimensionality directly. The mixB2DPPCA has two major advantages: 1) The model makes use of structured information of 2D data and can be easily extended for high order tensorial data. All the algorithm derivations remain without major difficulties. 2) mix2DPPCA carries over all the advantages of the mixture of PPCA.

The remainder of the paper is organized as follows. In Section \ref{Sec:II}, the mixture of bilateral-projection two-dimensional probabilistic PCA model is introduced. The variational approximation approach for solving the model is presented in Section \ref{Sec:III}. In Section \ref{Sec:IV}, some experimental results are conducted to evaluate the performance of the proposed model. Finally, conclusions are summarized in Section \ref{Sec:V}.

\section{Mixture of Bilateral-Projection 2DPPCA Model (mixB2DPPCA)}\label{Sec:II}
In this section, we introduce the mixture of bilateral-projection probabilistic 2DPCA model. For the purpose, we introduce several notations. Let $\mathcal X=\{\mathbf{X}_1,\mathbf{X}_2,...,\mathbf{X}_N\}$ be $N$ independent and identical random samples with values in $\mathbb{R}^{p\times q}$. For $n=1,...,N$, we suppose that sample $\mathbf X_n$ is generated independently from a mixture of $K$ underlying components with unknown probabilities $\pi_1,\pi_2,...,\pi_K$,
\begin{align}
p(\mathbf X_n|\mathbf B_n)= \sum_{k=1}^K\pi_k\mathcal N(\mathbf L_k\mathbf B_n^{(k)}\mathbf R_k^T+\mathbf M_k,\sigma_k\mathbf I,\sigma_k\mathbf I)\label{model}
\end{align}
where $\mathbf M_k\in\mathbb{R}^{p\times q}$ is the mean matrix, $\pi_k$s satisfy $\pi_k>0$ and $\sum_{k=1}^K \pi_k=1$, and $\mathbf{L}_k \in\mathbb{R}^{p\times r}$ and $\mathbf{R}_k\in\mathbb{R}^{q\times c}$ are the row and column loading matrices with $r\leq p,c\leq q$. Note that $\mathbf M_k$, $\mathbf L_k$ and $\mathbf R_k$ are associated with each component of mixture model, respectively. $\mathbf{B}_n^{(k)}\in\mathbb{R}^{r\times c}$ is the latent variable core of $\mathbf{X}_n$ associated with $k$-th matrix-variate Gaussian component \cite[Sec 3.3]{Timm2002} with $\sigma_k^2$ as residual variance.

Like \cite{Bishop2006}, we introduce a $K$-dimensional binary random variable $\mathbf z$ having a 1-of-$K$ representation in which a particular element $z_k$ is equal to 1 and all other elements are equal to 0. That is, $z_k\in\{0,1\}$ and $\sum_{k=1}^Kz_k=1$.
The distribution of $\mathbf z$ is defined by
\[
p(z_k=1):=\pi_k,
\]
which can be written as
\[
p(\mathbf z)=\prod_{k=1}^K\pi_k^{z_{k}}.
\]

Thus the conditional distribution of $\mathbf X_n$ given a particular value for $\mathbf z_n$ and $\mathbf B_n^{(k)}$ is the matrix-variate Gaussian
\begin{align*}
p(\mathbf X_n|z_{nk}=1,\mathbf B_n^{(k)})= \mathcal N(\mathbf X_n|\mathbf L_k\mathbf B_n^{(k)}\mathbf R_k^T+\mathbf M_k,\sigma_k\mathbf I,\sigma_k\mathbf I).
\end{align*}
Generally we have
\[
p(\mathbf X_n|\mathbf z_n,\mathbf B_n^{(k)})= \prod_{k=1}^K\mathcal N(\mathbf X_n|\mathbf L_k\mathbf B_n^{(k)}\mathbf R_k^T+\mathbf M_k,\sigma_k\mathbf I,\sigma_k\mathbf I)^{z_{nk}}
\]
In this model setting, the parameters are $\Theta = \{\pi_k,\mathbf M_k,\mathbf L_k,\mathbf R_k,\sigma^2_k\} (k=1,..,K)$, and the latent variables are $\mathbf z_n$ and $\mathbf B_n^{(k)}(n=1,...,N)$.


To develop a generative Bayesian model, we define a matrix-variate Gaussian prior $p(\mathbf B_n^{(k)})$ over the latent variable with zero-mean unit-covariance, defined as
\begin{equation*}
p(\mathbf B_n^{(k)})=\mathcal{N}(0,\mathbf{I}_r,\mathbf{I}_c)=\bigg(\frac{1}{2\pi}\bigg)^{\frac{rc}{2}}\cdot\exp\{-\frac{1}{2}\mathbf{tr}(\mathbf B_n^{(k)T}\mathbf B_n^{(k)})\}
\end{equation*}

Hence the joint log-likelihood of the observed  data set for such a mixture model is:
\begin{align*}
\mathcal L = \sum_{n=1}^N\sum_{k=1}^Kz_{nk}\ln\{\pi_kp(\mathbf X_n,\mathbf B_n^{(k)})\}.
\end{align*}

\section{Variational Approximation for mixB2DPPCA Model}\label{Sec:III}
We employ the Expectation Maximization (EM) algorithm to solve for model parameters $\Theta$. To maximize the log-likelihood of mixB2DPPCA, we take the expectation of $\mathcal L$ with respect to the posterior distribution of both $\mathbf B_n^{(k)}$ and $z_{nk}$, i.e.,
\begin{align}
\langle\mathcal L\rangle &= \sum_{n=1}^N\sum_{k=1}^K\langle z_{nk}\rangle\{\ln \pi_k-\frac{pq}{2}\ln\sigma^2_{k}-\frac{1}{2}\text{tr}(\langle\mathbf B_n^{(k)T}\mathbf B_n^{(k)}\rangle)\nonumber\\
&-\frac{1}{2\sigma^2_k}\text{tr}(\mathbf X_n-\mathbf M_k)^T(\mathbf X_n-\mathbf M_k)\nonumber\\
&+\frac{1}{\sigma^2}\text{tr}((\mathbf X_n-\mathbf M_k)^T\mathbf L_k \langle\mathbf B^{(k)}_n\rangle\mathbf R_k^T)\nonumber\\
&-\frac{1}{2\sigma^2}\text{tr}(\langle\mathbf B^{(k)T}_n\mathbf L_k^T\mathbf L_k \mathbf B^{(k)}_n\rangle\mathbf R_k^T\mathbf R_k)\label{L-fuction} 
\end{align}
where $\langle \cdot\rangle$ denotes the expectation.

In E-step, we update Q-distributions of all hidden variables $\mathbf B_n^{(k)}$ and $z_{n,k}$ with the current fixed parameter values for $\Theta$. In M-step, maximizing the function $\langle\mathcal L\rangle$ with respect to the model parameters $\Theta$, we can obtain `new' values for these parameters.

\subsection{\textbf{Variational E-step}}
\subsubsection{Update the Posterior Distribution of $z_{nk}$}
Suppose $\gamma_{nk}:=\langle z_{nk}\rangle$ and it is actually the posterior probability of $k$-mixture generating data point $\mathbf X_n$. By using the same strategy for the mixture Gaussian model \cite{Bishop2006}, we can obtain
\begin{align}
\gamma_{nk} = \frac{\pi_kp(\mathbf X_n|k)}{p(\mathbf X_n)}\label{gamma},
\end{align}
where $p(\mathbf X_n|k)$ is the $k$-the component, representing the marginal distribution for the observed data $\mathbf X_n$ over the latent variable. In our case, the marginal distribution of $\mathbf X_n$ is obtained by integrating out the latent variable $\mathbf B_n^{(k)}$:
\[
p(\mathbf X_n|k) = \int p(\mathbf X_n|\mathbf B_n^{(k)})p(\mathbf B_n^{(k)})d\mathbf B_n^{(k)}.
\]

Different from the vectorial PPCA, we note that the marginal distribution of the observed data $\mathbf X_n$ is in general no longer a matrix-variate Gaussian. Thus it is difficult to work with $p(\mathbf X_n|k)$ directly. Let $\mathbf {x}_n:=\text{vec}(\mathbf X_n)$, now we can work with $p(\mathbf x_n|k)$ instead of $p(\mathbf X_n|k)$. Fortunately, the marginal distribution of $\mathbf x_n$ is a multivariate Gaussian distribution when taking the special matrix-variate Gaussian prior $\mathbf B_n^{(k)}\sim\mathcal N(0,\mathbf I,\mathbf I)$. Let $\mathbf m_k = \text{vec}(\mathbf M_k)$
, we can obtain
\[
p(\mathbf x_n|k) \sim \mathcal N(\mathbf m_k,\mathbf C_k)
\]
where the observation covariance model is specified by $\mathbf C_k = (\mathbf R_k\mathbf R_k^T)\otimes(\mathbf L_k\mathbf L_k^T)+\sigma_k^2\mathbf I$. We refer readers to \cite{Bishop2006,Timm2002} for more details. Then the denominator in (\ref{gamma})
becomes
\begin{align*}
p(\mathbf x_n) = \sum_{k=1}^K\pi_k p(\mathbf x_n|k)
\end{align*}

After getting $\gamma_{nk}$, we update the estimated mean matrices $\mathbf M_k$'s and mixing proportions $\pi_k$'s, respectively, by
\begin{align}
\pi_k = \frac{1}{N}\sum_{n=1}^N\gamma_{nk}\quad \text{and} \quad\mathbf M_k = \frac{\sum_{n=1}^N \gamma_{nk}\mathbf X_n}{\sum_{n=1}^N\gamma_{nk}}\label{updatePi}
\end{align}

\subsubsection{Update the Posterior Distribution of $\mathbf B_n^{(k)}$}
 In computing the posterior distribution of $\mathbf B_n^{(k)}$, we encounter a difficulty that
the posteriori distribution of $\mathbf B_n^{(k)}$ given $\mathbf X_n$
\[
p(\mathbf B_n^{(k)}|\mathbf X_n,\mathbf L_k,\mathbf R_k,\sigma^2)\propto p(\mathbf X_n|\mathbf B_n^{(k)}, \mathbf L_k,\mathbf R_k,\sigma^2)p(\mathbf B_n^{(k)})
\]
is also in general not a matrix-variate Gaussian. To get a tractable posterior in the variational EM, we restrict the approximated variational distribution to be a matrix-variate Gaussian $\mathcal{N}(\mathbf{B}_n^{(k)}\,|\,\mathbf{Q}_n^{(k)}, \mathbf{T}_n^{(k)},\mathbf{S}_n^{(k)})$ to approximate the true posterior with the mean $\mathbf{Q}_n^{(k)}$ in size $r\times c$ and covariances $\mathbf{T}_n^{(k)}\succ 0$ of size $r\times r$  and $\mathbf{S}_n^{(k)} \succ 0$ of size $c\times c$, respectively. For mixB2DPPCA model, it follows as a natural extension of a single 2DPPCA. So the parameters $\mathbf Q_n^{(k)}$, $\mathbf T_n^{(k)}$ and $\mathbf S_n^{(k)}$ can be estimated through the maximization of a single likelihood function. Particularly, the derived formulas for estimating these parameters are given by, see more details in  \cite{YuBiYe2008},
\begin{align*}
\mathbf T_n^{(k)}=c\sigma_k^2[\text{tr}(\mathbf R^T_k\mathbf R_k\mathbf S_n^{(k)})\mathbf L_k^T\mathbf L_k+\sigma_k^2\text{tr}(\mathbf S_n^{(k)})\mathbf I_r]^{-1}
\end{align*}
\begin{align*}
\mathbf S_n^{(k)}=r\sigma_k^2[\text{tr}(\mathbf L_k^T\mathbf L_k\mathbf T_n^{(k)})\mathbf R^T_k\mathbf R_k+\sigma_k^2\text{tr}(\mathbf T_n^{(k)})\mathbf I_c]^{-1}
\end{align*}
and each $\mathbf Q_n^{(k)}$ needs to satisfy
\[
\mathbf L^T_k\mathbf L_k\mathbf Q_n^{(k)}\mathbf R_k^T\mathbf R_k+\sigma_k^2\mathbf Q_n^{(k)}=\mathbf L^T_k(\mathbf X_n-\mathbf M_k)\mathbf R_k.
\]
To solve this we need to make a vectorization on both sides and solve a linear equation
\begin{align}
(\mathbf R_k^T\mathbf R_k\otimes\mathbf L^T_k\mathbf L_k+\sigma_k\mathbf I\otimes \sigma_k\mathbf I)\text{vec}(\mathbf Q^{(k)}_n)=\mathbf y^{(k)}_n\label{updateQ}
\end{align}
with respect to $\text{vec}(\mathbf Q_n^{(k)})$, where
\begin{align*}
\mathbf y^{(k)}_n=\text{vec}(\mathbf L^T_k(\mathbf X_n-\mathbf M_k)\mathbf R_k)
\end{align*}
then reshape $\text{vec}(\mathbf Q_n^{(k)})$ back to get $\mathbf Q_n^{(k)}$.

As we assume the approximated posterior distribution of $\mathbf B_n^{(k)}$ is matrix-variate Gaussian, so we can get $\langle\mathbf B_n^{(k)}\rangle = \mathbf Q_n^{(k)}$  and the following second-order expectations:
\begin{align}
&\langle\mathbf B_n^{(k)T}\mathbf B_n^{(k)}\rangle = \mathbf Q_n^{(k)T}\mathbf Q_n^{(k)}+\mathbf S_n^{(k)}\text{tr}(\mathbf T_n^{(k)})\label{updateBB}\\
&\langle\mathbf B^{(k)T}_n\mathbf L_k^T\mathbf L_k \mathbf B^{(k)}_n\rangle=\mathbf Q^{(k)T}_n\mathbf L_k^T\mathbf L_k \mathbf Q^{(k)}_n+\mathbf S_n^{(k)}\text{tr}(\mathbf T_n^{(k)}\mathbf L_k^T\mathbf L_k)\label{updateBLLB}
\\
&\langle\mathbf B^{(k)}_n\mathbf R_k^T\mathbf R_k\mathbf B^{(k)T}_n\rangle=\mathbf Q^{(k)}_n\mathbf R_k^T\mathbf R_k\mathbf Q^{(k)T}_n+\mathbf T_n^{(k)}\text{tr}(\mathbf S_n^{(k)}\mathbf R_k^T\mathbf R_k)\label{updateBRRB}
\end{align}

\subsection{\textbf{Variational M-step}}
In the M-step, we fix all the distributions over the hidden variables and gather all the terms containing parameters $\mathbf L_k$, $\mathbf R_k$ and $\sigma_k^2$ in (\ref{L-fuction}) to
maximize them respectively. It turns out that:
\begin{align}
\mathbf L_k=&[\sum_{n=1}^N\gamma_{nk}(\mathbf X_n-\mathbf M_k)\mathbf R_k\langle\mathbf B_n^{(k)}\rangle^T]\nonumber\\
&\times[\sum_{n=1}^N\gamma_{nk}\langle \mathbf B_n^{(k)}\mathbf R^T_k\mathbf R_k\mathbf B_n^{(k)T}\rangle]^{-1}\label{updateL}
\end{align}
\begin{align}
\mathbf R_k=&[\sum_{n=1}^N\gamma_{nk}(\mathbf X_n-\mathbf M_k)^T\mathbf L_k\langle\mathbf B_n^{(k)}\rangle]\nonumber\\
&\times[\sum_{n=1}^N\gamma_{nk}\mathbf \langle \mathbf B_n^{(k)T}\mathbf L^T_k\mathbf L_k\mathbf B_n^{(k)}\rangle]^{-1}\label{updateR}
\end{align}
and
\begin{align}
\sigma^2_k &= \frac{1}{pqN_k}\{\sum_{n=1}^N\gamma_{nk}\text{tr}(\mathbf X_n-\mathbf M_k)^T
(\mathbf X_n-\mathbf M_k)\nonumber\\
&-2\sum_{n=1}^N\gamma_{nk}\text{tr}(\mathbf R_k\langle\mathbf B^{(k)}_n\rangle^T\mathbf L_k^T(\mathbf X_n-\mathbf M_k))\nonumber\\
&+\sum_{n=1}^N\gamma_{nk}\text{tr}(\langle\mathbf B^{(k)T}_n\mathbf L_k^T\mathbf L_k\mathbf B^{(k)}_n\rangle\mathbf R_k^T\mathbf R_k)\}\label{updateSigma}
\end{align}
where $N_k=\sum_n\gamma_{nk}$.

The overall variational EM algorithm is to alternate between E-step and M-step.
 The final variational EM algorithm is summarized
in Algorithm \ref{alg1}.
\begin{algorithm}
\renewcommand{\algorithmicrequire}{\textbf{Initialize:}}
\renewcommand\algorithmicensure {\textbf{Variational E-step:} }
\renewcommand\algorithmicensure {\textbf{Variational M-step:} }
\caption{Variational EM algorithm for mixB2DPPCA.}\label{alg1}
\begin{algorithmic}[1]
  \REQUIRE Training set $\mathcal{X} = \{\mathbf X_n\}_{n=1}^N$; Initialize all of model parameters $\Theta$ and covariance matrices $\mathbf T_n^{(k)}$ and $\mathbf S_n^{(k)}$, $n=1,...,N$ and $k=1,...,K$.
 \FOR {$t=1$ to $T$}
    \STATE \textbf{Variational E-step:}
    \begin{itemize}
             \item Iterate the mean matrix $\mathbf Q_n^{(k)}$ based on (\ref{updateQ}) and update the second-order expectations based on (\ref{updateBB}), (\ref{updateBLLB}) and (\ref{updateBRRB}).\\
           \end{itemize}
       \begin{itemize}
           \item  Update each $\gamma_{nk}$, mixing proportions $\pi_k$ and mean matrices $\mathbf M_k$ based on (\ref{gamma}) and  (\ref{updatePi}).
           \end{itemize}

     \STATE \textbf{Variational M-step:}
            \begin{itemize}
           \item   Maximize objective function  $\langle\mathcal L\rangle$ with respect to each elements $\mathbf L_k$, $\mathbf R_k$ and $\sigma_k^2$ based on (\ref{updateL}), (\ref{updateR}) and (\ref{updateSigma}).
           \end{itemize}
 \ENDFOR
\end{algorithmic}
\end{algorithm}

Define the average reconstruction error
\begin{align}
\mathbf e(t) =\sqrt{\frac{\sum_{n=1}^N\|\mathbf X_{n}-\widehat{\mathbf X}_{n}^{(t)}\|_F^2}{N}}\label{comput_error}
\end{align}
where $\widehat{\mathbf X}_n = \mathbf L_{k'}\mathbf B^{(k')}_n \mathbf R^T_{k'} + \mathbf M_{k'}$ with $k'=\text{arg}\max_k\{\gamma_{nk}\}$ the reconstructed image.

Algorithm \ref{alg1} may terminate either a given maximum iterative number $T$ is achieved or the following condition is satisfied,
\begin{align}
|\mathbf e(t)-\mathbf e(t+1)|\leq \epsilon \label{epsilon}
\end{align}
where $\epsilon$ is a given error tolerance.

\subsection{The Reduced-Dimensionality Representation for a New Sample}
In order to obtain the reduced-dimensionality representation for a given sample, we should solve for the latent variable cores. From the probabilistic perspective, the posterior mean $\mathbf Q_{new}^{(k)}:=\langle\mathbf B_{new}^{(k)}|\mathbf X_{new}\rangle$ can be seen as the reduced-dimensionality representation, which is a $r\times c$ feature matrix and given by solving a linear equation
\begin{align*}
(\mathbf R_k^T\mathbf R_k\otimes\mathbf L^T_k\mathbf L_k+\sigma_k\mathbf I\otimes \sigma_k\mathbf I)\text{vec}(\mathbf Q^{(k)}_{new})=\mathbf y_{new}^{(k)}
\end{align*}
with respect to $\text{vec}(\mathbf Q_{new}^{(k)})$, where
\begin{align*}
\mathbf y_{new}^{(k)}=\text{vec}(\mathbf L^T_k(\mathbf X_{new}-\mathbf M_k)\mathbf R_k)
\end{align*}
then reshape $\text{vec}(\mathbf Q_{new}^{(k)})$ back to get $\mathbf Q_{new}^{(k)}$. As the same time, we can compute the corresponding $\gamma_{new,k}$, i.e., the posterior probability of $k$-th component generating the new sample, given by
\[
\gamma_{new,k} = \frac{p(\mathbf X_{new}|k)\pi_k}{p(\mathbf X_{new})}
\]

We find the largest $\gamma_{new,k}$ ($k=1,...,K$) from which the most appropriate local 2DPPCA model can be identified for the new sample. That is, a natural choice is to assign the new sample to a cluster with the largest posterior probability.
  \section{Experimental Results and Analysis}\label{Sec:IV}
In this section, we conduct several experiments on some public databases to assess the proposed mixB2DPPCA model. These experiments are designed to evaluate the performance of the proposed mix2DPPCA in reconstruction and recognition by comparing with existing models and algorithms. 

The relevant PCA algorithms that can be fairly compared against our proposed mixB2DPPCA are GLRAM (Generalized Low Rank Approximations of Matrices) \cite{Ye2005}, PSOPCA (Probabilistic Second-Order PCA) \cite{YuBiYe2011}, mixture of PPCA \cite{TippingBishop1999a} with the code from {\small\url{http://www.science.uva.nl/\~jverbeek}}. 
Because the zero-noise PSOPCA model and GLRAM have the same stationary point \cite{YuBiYe2011}, we only compare with GLRAM.
\subsection{Data Preparation and Experiment Setting}\label{sectionIVA}
All of the experiments are conducted on the following four public available datasets:
\begin{itemize}
  \item A subset of handwritten digits images from the MNIST database ({\small\url{http://yann.lecun.com/exdb/mnist}}).
  \item The Yale face database ({\small\url{http://vision.ucsd.edu/content/yale-face-database}}).
  \item The AR face database ({\small\url{http://rvl1.ecn.purdue.edu/aleix/aleix\_face\_DB.html}}).
    \item The FERET face database ({\small\url{http://www.itl.nist.gov/iad/humanid/feret/feret\_master.html}}).
\end{itemize}

The subset of handwritten digits images is selected from MNIST database, which contains 1000 digital images with 100 images of each digit. All images are in grayscale and have a uniform size of $28\times 28$ pixels.

The Yale face database contains 15 individuals, with 11 images for each individual. The images were captured under different illumination and expression conditions. 
The images are all $100\times 100$ pixels with 256 grey levels. In the experiments, we randomly select 6 images of each person as the training samples, and use the remaining images to form the testing sample set. All images are scaled to a resolution of $64\times 64$ pixels.

\begin{figure*}
\begin{center}
   {\includegraphics[width=175mm,height=45mm]{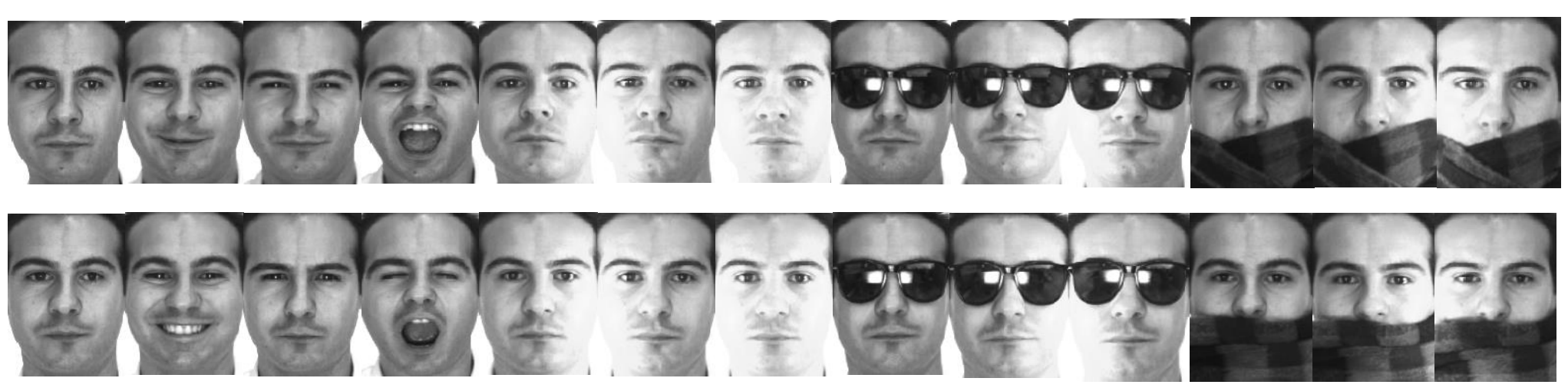}}
\end{center}
\caption{Twenty-six face examples of one subject from AR database. The first row is from the first session, and the second row images are from the second session.}\label{ARdatabase}
\end{figure*}

The AR face database contains over 4,000 color images corresponding to 126 subjects. There are variations of facial expressions, illumination conditions, and occlusions (sun glasses and scarf) with each person. Each individual consists of 26 frontal view images taken in two sessions (separated by 2 weeks), where each session has 13 images. Figure \ref{ARdatabase} shows the 26 images of one subject. In the experiments, we select 30 subjects (15 man and 15 women), and only use the non-occluded 14 images (i.e., the first seven face images of each row in Figure \ref{ARdatabase}). The first seven of each subject are used for training and the last seven for testing. All images are cropped and resized to $50\times 40$ pixels.

FERET database includes 1400 images of 200 different subjects, with
7 images per subject. In the experiments, we select 50 subjects randomly. Five images of each subject are used for training and the remained images are used for testing. All images are cropped and resized to $32\times 32$ pixels.

In experiments, the initial mixing proportions are set to $\pi_k = 1/K$ and the initial loading matrices $\mathbf L_k$ and $\mathbf R_k$ are given randomly. Besides, we choose randomly $K$ samples as mean matrices $\mathbf M_k$ of the mixture gaussian model and set all $\sigma_k^2=1$.
\subsection{Reconstruction Performance}
In this section, we test reconstruction error of the proposed mixB2DPPCA model (\ref{model}). Applying the proposed model, all digital images can be softly grouped into $K$ clusters,  each of which is modelled by a local B2DPPCA. From all the trained $\gamma_{nk}$, the most appropriate local B2DPPCA for a given sample can be found. Then we use the most appropriate local B2DPPCA to reconstruct the initial digit image, that is:
\[
\widehat{\mathbf X}_n =  \mathbf L_{k'}*\mathbf{Q}_n^{(k')}*\mathbf R_{k'}^T + \mathbf{M}_{k'}.
\]
where $k^{'}$ represents the $k^{'}$-th local B2DPPCA which most appropriate to the sample $\mathbf X_n$. After obtaining all reconstructed digit images $\widehat{\mathbf X}_n$, we can using the equation (\ref{comput_error}) to compute the average reconstruction error.

Next we compare the reconstruction error of different algorithms on three databases. 
In all algorithms, we set the iterative number is  $T = 50$ and the reduced dimension is $r=c=4$.
\subsubsection{Reconstruction Error on Digit Image Set}
We use the given digital image subset in Section \ref{sectionIVA}  as training set. In this phase, we compare the reconstruction error of the training set.
\begin{figure*}
\begin{center}
   {\includegraphics[width=50mm,height=40mm]{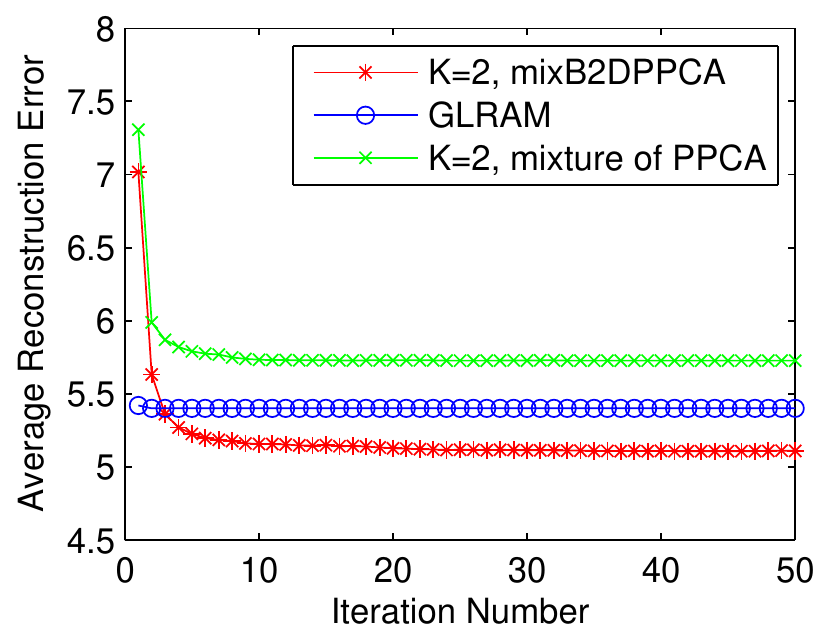}}
   {\includegraphics[width=50mm,height=40mm]{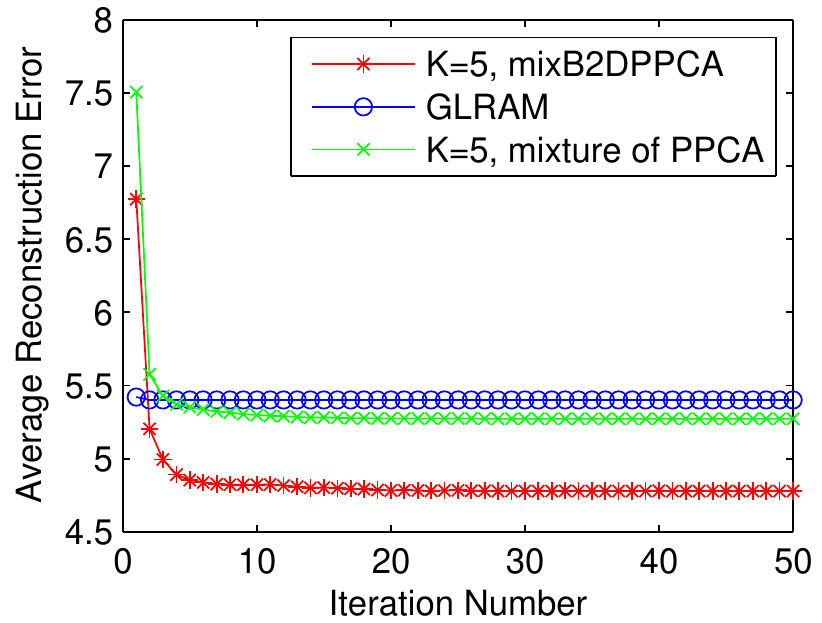}}
   {\includegraphics[width=50mm,height=40mm]{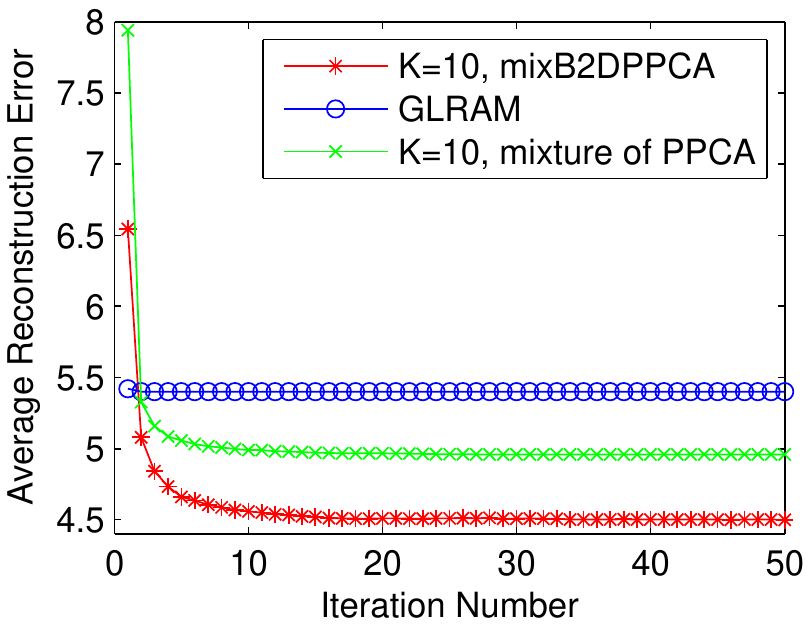}}
\end{center}
\caption{Average reconstruction error versus iteration number with the components number $K=2$, $K=5$ and $K=10$ from the left to right.}\label{err_digit}
\end{figure*}

Figure \ref{err_digit} shows the average reconstruction error of the relevant algorithms. From left to right, the component number is $K=2$, $K=5$ and $K=10$ respectively. Firstly, from these three sub-figures, we can see that the reconstruction error of GLRAM algorithm has no change. This is because GLRAM has no relationship with $K$. Besides, GLRAM works by iteratively computing the leading eigenvectors of the left and right one-sided sample covariance matrices. Thus GLRAM convergent in five steps and the change of reconstruction error is not obvious in the figure. Secondly, fixing the same number of reduced dimension, the performance of our proposed mixB2DPPCA is better than GLRAM. From the view of compression, decoded images from our algorithm have higher quality for the compression ratio of $49:1$. It illustrates that mixB2DPPCA can correctly identify data according to clusters. When $K$ becomes larger, the mixB2DPPCA outperform  the mixture of PPCA in terms of reconstruction errors.
\begin{figure}
\begin{center}
   {\includegraphics[width=60mm,height=60mm]{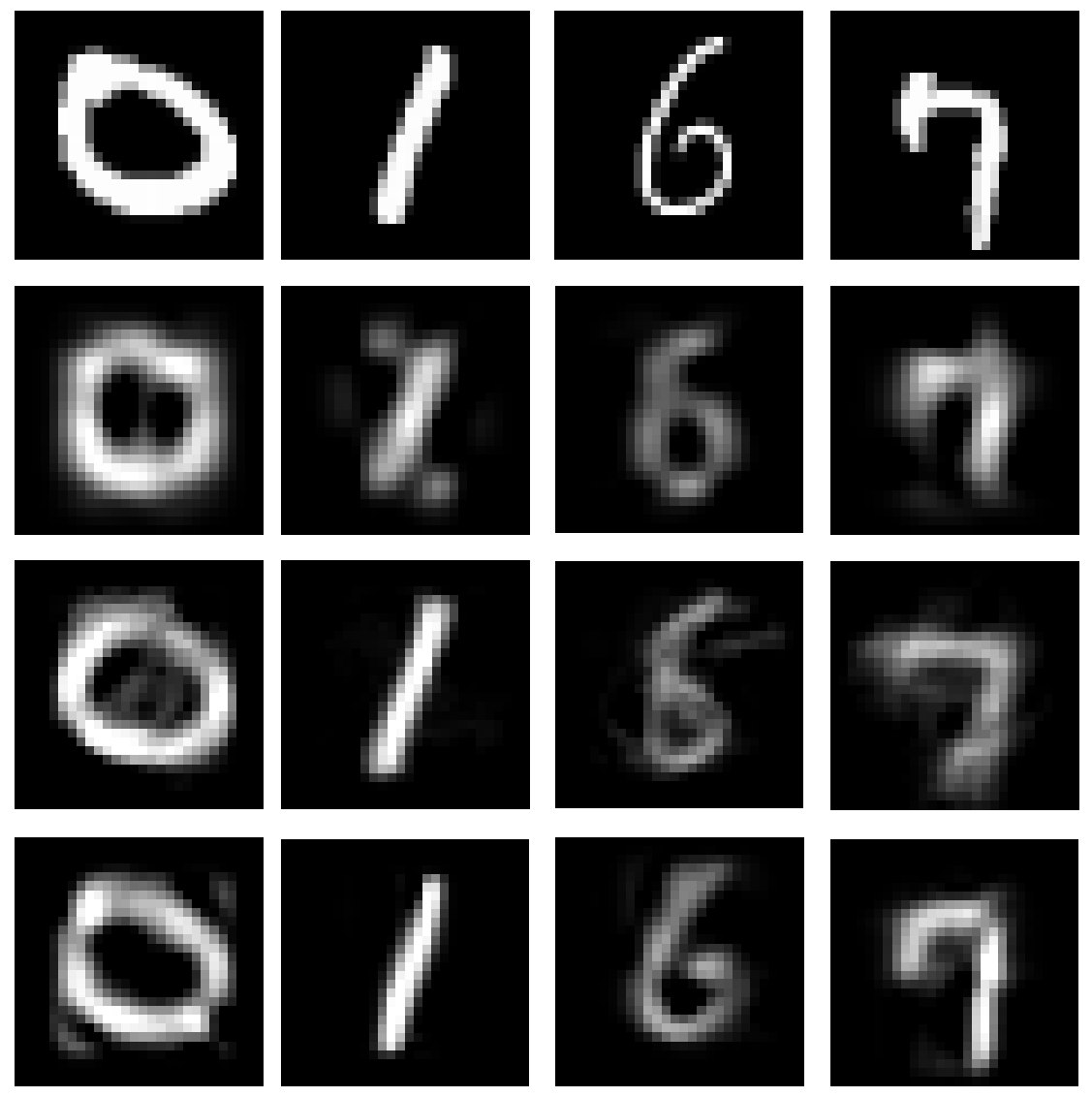}}
\end{center}
\caption{Original images and reconstructed images: The first row shows four original digital images. The second, third and fourth rows are the reconstructed images by GLRAM, mixture of PPCA and mixB2DPPCA, respectively.}\label{cons_digit}
\end{figure}

The reconstructed images of different methods are shown in Fig. \ref{cons_digit} with $K = 10$. The first row shows three original images. The second, third and fourth rows are the reconstructed images by GLRAM, mixture of PPCA and mixB2DPPCA, respectively. It can be found that the proposed mixB2DPPCA has better reconstruction outcomes, while the results of other two methods show a litter degradation.
\begin{table*}
\renewcommand\arraystretch{1.0}
  \centering
  \footnotesize
   \begin{tabular}{|c|c|r|c|r|c|r|c|}
      \hline
      \multirow{2}*{$r,c$}  &\multirow{2}*{GLRAM}   &\multicolumn{2}{c|}{$K=4$} &\multicolumn{2}{c|}{$K=6$} &\multicolumn{2}{c|}{$K=8$}  \\
 \cline{3-8}
     & &  mixPPCA& mixB2DPPCA&  mixPPCA& mixB2DPPCA&  mixPPCA& mixB2DPPCA \\
     \hline
     2   & 0.6133 & 0.5760$\pm$0.0401 & 0.6400$\pm$0.0267& 0.6237$\pm$0.0219 &\textbf{0.6720}$\pm$\textbf{0.0210} &0.6519$\pm$0.0124&0.6693$\pm$0.0250   \\
\hline
      4       & 0.7067&0.6376$\pm$0.0222  & 0.7173$\pm$0.0197& 0.6642$\pm$0.0245 &0.7146$\pm$0.0180&0.6613$\pm$0.0201 &\textbf{0.7200}$\pm$\textbf{0.0089}  \\
\hline
       6       & 0.7200& 0.6480$\pm$0.0289& 0.7200$\pm$0.0154& 0.6506$\pm$0.0186& 0.7187$\pm$0.0203&0.6480$\pm$0.0283&\textbf{0.7320}$\pm$\textbf{0.0160}  \\
 \hline
       8       &0.7200 & 0.6786$\pm$0.0117& 0.7240$\pm$0.0227&0.6560$\pm$0.0265  &0.7187$\pm$0.0262 &0.6640$\pm$0.0233& \textbf{0.7347}$\pm$\textbf{0.0160} \\
     \hline
   \end{tabular}
  \caption{Recognition accuracy of GLRAM, mixture of PPCA and mixB2DPPCA training on the Yale database}\label{Table_Yale}
\end{table*}
\begin{table*}
\renewcommand\arraystretch{1.0}
  \centering
  \footnotesize
   \begin{tabular}{|c|c|r|c|r|c|r|c|}
      \hline
     \multirow{2}*{$r,c$}  &\multirow{2}*{GLRAM}   &\multicolumn{2}{c|}{$K=6$} &\multicolumn{2}{c|}{$K=8$} &\multicolumn{2}{c|}{$K=10$}  \\
 \cline{3-8}
     & &  mixPPCA& mixB2DPPCA&  mixPPCA& mixB2DPPCA&  mixPPCA& mixB2DPPCA \\
     \hline
      4       & 0.5714& 0.5328$\pm$0.0220 &0.6671$\pm$0.0333 & 0.5595$\pm$0.0214 & 0.7000$\pm$0.0371&0.5752$\pm$0.0297& \textbf{0.7244}$\pm$\textbf{0.0381}  \\
\hline
       6       & 0.6857& 0.6252$\pm$0.0242&0.7867$\pm$0.0138 & 0.6343$\pm$0.0236 & \textbf{0.8017}$\pm$\textbf{0.0291}&0.6613$\pm$0.0182&0.7576$\pm$0.0366   \\
 \hline
       8       &0.7190 & 0.7004$\pm$0.0190& 0.8116$\pm$0.0231 & 0.7100$\pm$0.0246 &0.8211$\pm$0.0246 &0.7133$\pm$0.0222& \textbf{0.8357}$\pm$\textbf{0.0237}  \\
     \hline
   \end{tabular}
  \caption{Recognition accuracy of GLRAM, mixture of PPCA and mixB2DPPCA training on the AR database}\label{Table_AR}
\end{table*}
\begin{table*}
\renewcommand\arraystretch{1.0}
  \centering
  \footnotesize
   \begin{tabular}{|c|c|r|c|r|c|r|c|}
      \hline
     \multirow{2}*{$r,c$}  &\multirow{2}*{GLRAM}   &\multicolumn{2}{c|}{$K=6$} &\multicolumn{2}{c|}{$K=8$} &\multicolumn{2}{c|}{$K=10$}  \\
 \cline{3-8}
     & &  mixPPCA& mixB2DPPCA&  mixPPCA& mixB2DPPCA&  mixPPCA& mixB2DPPCA \\
     \hline
      4       & 0.5000& 0.4620$\pm$0.0315 &0.6070$\pm$0.0427 & 0.4690$\pm$0.0470 & \textbf{0.6210}$\pm$\textbf{0.0326}&0.4840$\pm$0.0316& 0.5900$\pm$0.0429  \\
\hline
       6       & 0.5300& 0.5140$\pm$0.0206&\textbf{0.6733}$\pm$\textbf{0.0541} & 0.5350$\pm$0.0283 & 0.6467$\pm$0.0343&0.5320$\pm$0.0297&0.6644$\pm$0.0328   \\
 \hline
       8       &0.5400 & 0.5440$\pm$0.0298& 0.6900$\pm$0.0458 & 0.5610$\pm$0.0159 &\textbf{0.6945}$\pm$\textbf{0.0526} &0.5580$\pm$0.0187& 0.6770$\pm$0.0593  \\
     \hline
     10      &0.5500 & 0.5910$\pm$0.0460& 0.6890$\pm$0.0455 & 0.5720$\pm$0.0364 &0.6960$\pm$0.0599 &0.5970$\pm$0.0336& \textbf{0.7100}$\pm$\textbf{0.0573}  \\
     \hline
   \end{tabular}
  \caption{Recognition accuracy of GLRAM, mixture of PPCA and mixB2DPPCA training on the FERET database}\label{Table_FERET}
\end{table*}
\subsubsection{Reconstruction Error on Yale and AR Databases}
In this experiment, we compare the reconstruction error on Yale and AR databases.
\begin{figure}
\begin{center}
   \subfloat[]{\includegraphics[width=41mm,height=36mm]{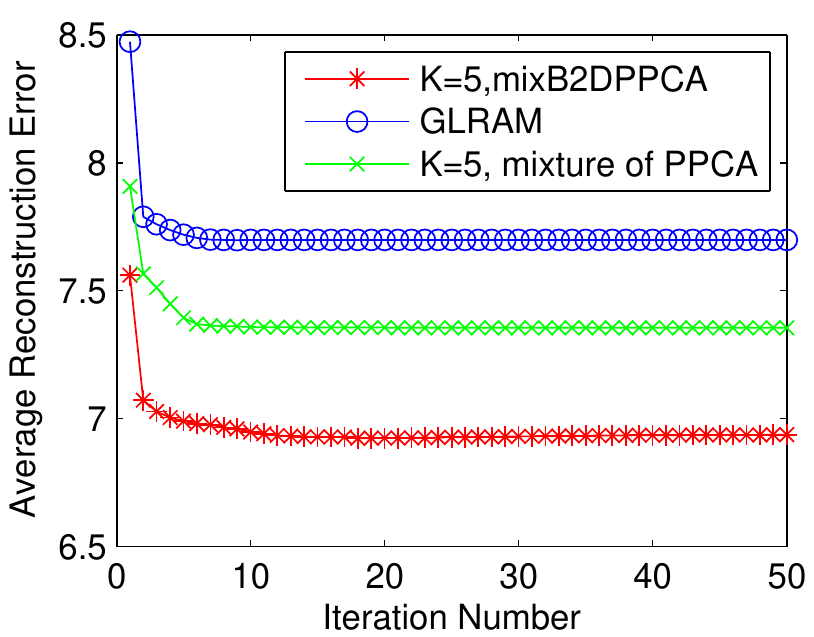}}
   \subfloat[]{\includegraphics[width=41mm,height=36mm]{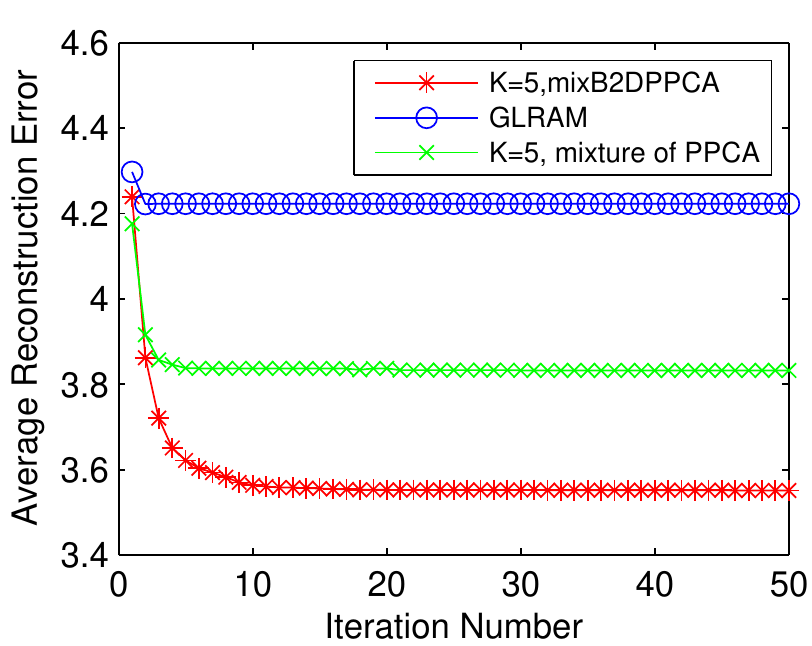}}
\end{center}
\caption{Average reconstruction error versus iteration number with the components number $K=5$ on Yale database (a) and AR database (b).}\label{err_Yale}
\end{figure}
Figure \ref{err_Yale} shows the average reconstruction error of all the algorithms: (a) on the Yale database and (b) on the AR database. The component number is $K=5$ and the reduced dimensionality is $(r,c) =(4,4)$. It is obvious that the reconstruction error of mixB2DPPCA on testing set has reduced greatly than other algorithms.

\begin{figure}
\begin{center}   {\includegraphics[width=65mm,height=65mm]{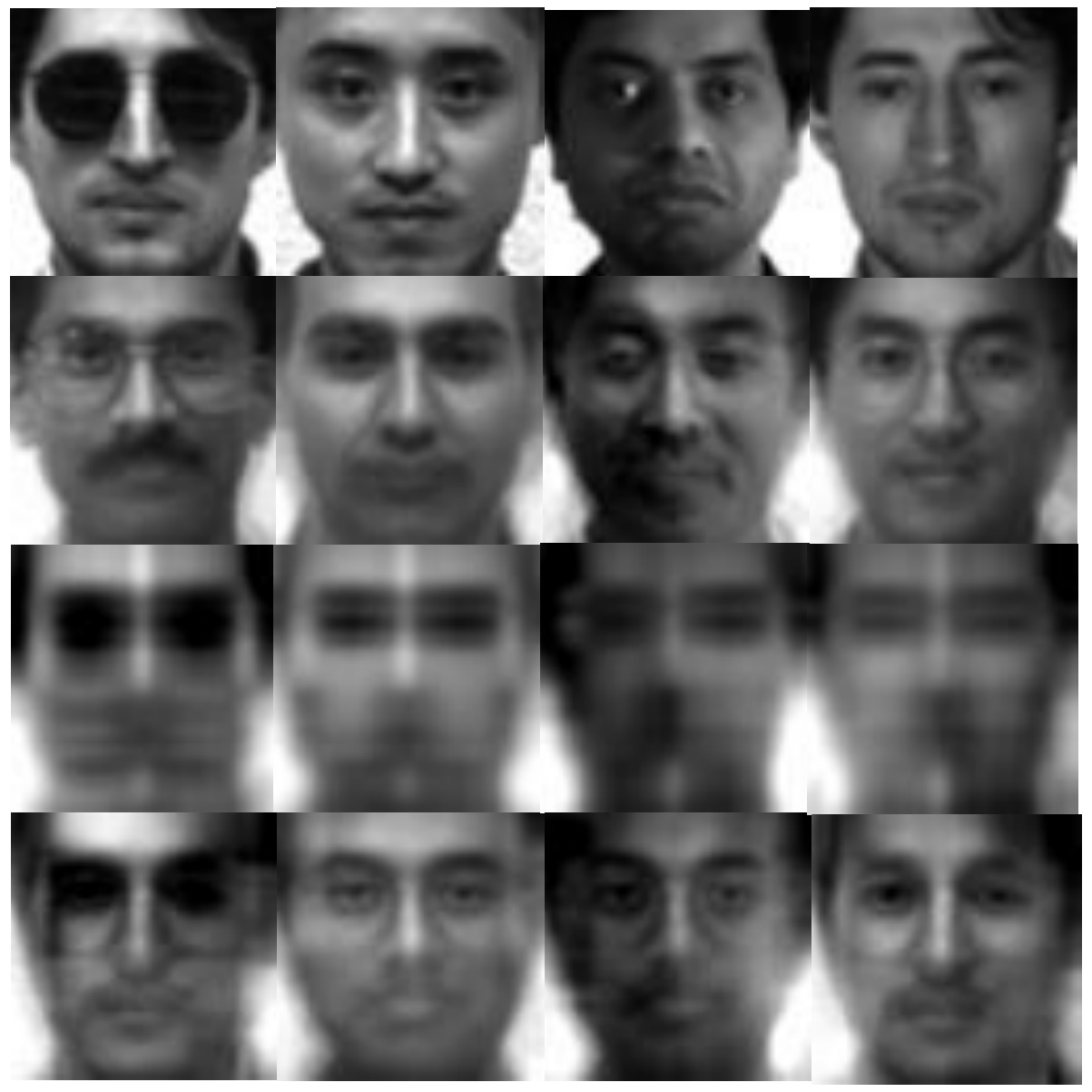}}
\end{center}
\caption{Original images (in the Yale database) and reconstructed images: The first row is original images. The second, third and fourth rows are the reconstructed images by mixture PPCA, GLRAM and mixB2DPPCA, repectively.}\label{Cons_Yale}
\end{figure}

\begin{figure}
\begin{center}
   {\includegraphics[width=80mm,height=68mm]{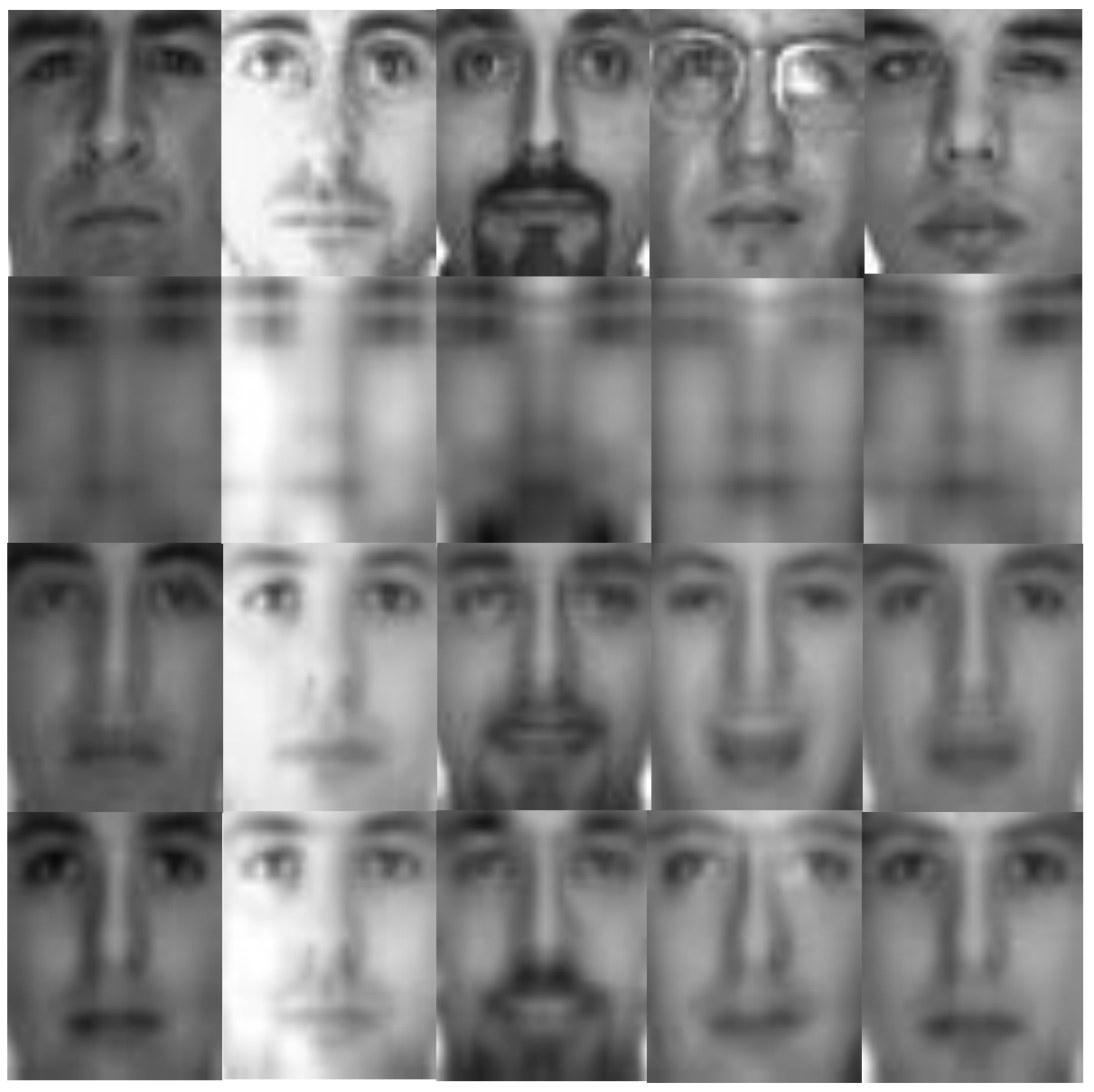}}
\end{center}
\caption{Original images (in the AR database) and reconstructed images: The first row is the original images. The second, third and fourth rows are the reconstructed images by GLRAM, mixture PPCA and mixB2DPPCA, repectively.}\label{Cons_AR}
\end{figure}

Figure \ref{Cons_Yale} shows some reconstructed images of different algorithms on Yale database. The first row is four original images. The second, third, and fourth rows are the corresponding images reconstructed by mixture of PPCA, GLRAM and mixB2DPPCA. It can be shown that the results of our algorithm have better visual effect than that of GLRAM. Besides we can also see that although the face images reconstructed by mixture of PPCA are relatively clear, they don't match the same original images visually. The reconstructed images on AR database are shown in Figure \ref{Cons_AR}. The first row shows five original images in the test set and the last three rows are the reconstructed images from three models.

From the reconstruction experiments, we can conclude that the mixB2DPPCA generally outperforms global linear 2DPCA algorithms in terms of reconstruction errors. It demonstrates that the classification of training set in advanced is important for the performance of feature extraction.
\subsection{Recognition Performance}
In this section, we compare the recognition performances of GLRAM, mixture of PPCA and mixB2DPCA on Yale, AR and FERET face databases. These algorithms can be used for extracting features of facial images from the training samples, respectively, and then a nearest neighbor classifier (1-NN) is used to find the most-similar face from the training samples for a querying face. In our experiments, the distance measure between two sets of feature matrices $\mathbf B_{n_1}$ and $\mathbf B_{n_2}$, is defined as
\[
\text{dist} = \sum^K_{k=1} \|\mathbf B^{(k)}_{n_1}-\mathbf B^{(k)}_{n_2}\|_F.
\]
where $\mathbf B_{n}=[\mathbf B_{n}^{(1)},\mathbf B_{n}^{(2)},...,\mathbf B_{n}^{(K)}]$ represents the combination of $K$ latent variable cores related with $n$-th sample\footnote{A more accurate way is to use $\gamma_{n_1k}\gamma_{n_2k}$ to weight the individual distance.}. In all algorithms, we set maximum iteration number is 50 and $\epsilon$ in (\ref{epsilon}) is 1E-3. We repeat the procedure 10 times, and the mean values and relevant variances are reported in Tables \ref{Table_Yale} to \ref{Table_FERET}.

Table \ref{Table_Yale} shows the recognition rates of three feature extraction algorithms: GLRAM, mixture of PPCA and mixB2DPPCA training on Yale database. The mean values and relevant variances are reported for the cases of the reduced dimension $(r,c)=(2,2)$, $(4,4)$, $(6,6)$ and $(8,8)$. For the mixture of PPCA and mixB2DPPCA, we also computed the recognition rates for the different component number $K$ ($K=4, 6, 8$), shown in Table \ref{Table_Yale}. Firstly, from the table we can see that the recognition rates of the mixture of PPCA and mixB2DPPCA have a little fluctuation compared with GLRAM. This may be caused by the uncertainty of probability. Secondly, compared with GLRAM, the mean recognition rates of mixB2DPPCA algorithm have obviously improved. The bold figures are the best results in the comparison.

Table \ref{Table_AR} shows the recognition rates of the above three algorithms training on AR database. The reduced dimensions are $(r,c)=(4,4)$, $(6,6)$ and $(8,8)$ and component numbers are $K=6, 8, 10$, respectively. From the table we can see that the mean recognition rates of mixB2DPPCA algorithm have better improvement over the other two algorithms. 

Table \ref{Table_FERET} shows the recognition rates on FERET database. The reduced dimensions are $(r,c)=(4,4)$, $(6,6)$, $(8,8)$ and $(10,10)$, and the component numbers are $K=6, 8, 10$,  respectively. In this case, both the mixture of PPCA and the proposed mixB2DPPCA produce slightly larger variances, however the mean recognition rates have risen greatly. GLRAM is relatively more robust.

\section{Conclusions}\label{Sec:V}
In this paper, we proposed a mixture of bilateral-projection probabilistic PCA model for feature extraction and dimensionality reduction for 2D data. Different from the standard PCA which is a global dimension reduction model, this model employs the mixture of matrix-variate Gaussian to model local linear sub-models. All the parameters in the resulting probabilistic model can be estimated through the maximization of the likelihood function. The new model not only makes good use of spatial (structural) information of 2D data but also can softly group data into a given number of clusters. The performance of feature extraction of the proposed method generally outperforms other existing 2D  algorithms in terms of reconstruction error and recognition rate. The approach used in this paper can be readily extended to higher order tensorial data and other non-Gaussian noise models can also be integrated into the model such.


\end{document}